\documentclass[pdflatex,sn-mathphys-num]{sn-jnl}

\usepackage{graphicx}%
\usepackage{multirow}%
\usepackage{amsmath,amssymb,amsfonts}%
\usepackage{amsthm}%
\usepackage{mathrsfs}%
\usepackage[title]{appendix}%
\usepackage{xcolor}%
\usepackage{textcomp}%
\usepackage{manyfoot}%
\usepackage{booktabs}%
\usepackage{caption}
\usepackage{placeins}
\usepackage{subcaption}
\usepackage{algorithm}%
\usepackage{algorithmicx}%
\usepackage{algpseudocode}%
\usepackage{listings}%
\usepackage{soul}
\usepackage{array}
\newcolumntype{L}[1]{>{\raggedright\arraybackslash}p{#1}}
\newcolumntype{Y}{>{\raggedright\arraybackslash}X}
\usepackage{booktabs}
\usepackage{tabularx}
\usepackage{float}
\newcommand{\eightpt}{\fontsize{8pt}{8pt}\selectfont}

\theoremstyle{thmstyleone}%
\theoremstyle{thmstyletwo}%

\theoremstyle{thmstylethree}%
\usepackage[most]{tcolorbox}
\definecolor{myblockbg}{rgb}{0.94,0.94,1} 
\definecolor{myblocktitle}{rgb}{0.2,0.2,0.6} 

\newtcolorbox{customblock}[1][]{
  colback=myblockbg,
  colframe=myblocktitle,
  coltitle=myblocktitle,
  fonttitle=\bfseries,
  enhanced,
  sharp corners=south,
  boxrule=0.8pt,
  title=#1
}

\makeatletter
\renewcommand{\equalcont}[1]{%
  \gdef\@equalconttext{#1}%
  \global\equalconttrue%
}
\makeatother
\begin{document}

\title[AI for rare disease phenotyping from clinical notes]{An artificial intelligence framework for end-to-end rare disease phenotyping from clinical notes using large language models}


\author*[1,2,3]{\fnm{Cathy} \sur{Shyr}}\email{cathy.shyr@vumc.org}

\author[4]{\fnm{Yan} \sur{Hu}}\email{yanhu.yhu@gmail.com}

\author[5]{\fnm{Rory J.}\sur{Tinker}}\email{rory.tinker@mssm.edu}

\author[3]{\fnm{Thomas A.}\sur{Cassini}}\email{thomas.a.cassini@vumc.org}

\author[6]{\fnm{Kevin W.}\sur{Byram}}\email{kevin.byram@vumc.org}

\author[3]{\fnm{Rizwan}\sur{Hamid}}\email{rizwan.hamid@vumc.org}

\author[1,7]{\fnm{Daniel V.}\sur{Fabbri}}\email{daniel.fabbri@vumc.org}

\author[1,6]{\fnm{Adam}\sur{Wright}}\email{adam.wright@vumc.org}

\author[1,6]
{\fnm{Josh F.}\sur{Peterson}}\email{josh.peterson@vumc.org}

\author[1]
{\fnm{Lisa}\sur{Bastarache}\textsuperscript{\ensuremath{\dagger}}}\email{lisa.bastarache@vumc.org}

\author[8]{\fnm{Hua} \sur{Xu}\textsuperscript{\ensuremath{\dagger}}}\email{hua.xu@yale.edu}

\affil*[1]{\orgdiv{Department of Biomedical Informatics}, \orgname{Vanderbilt University Medical Center}, \orgaddress{\street{2525 West End Avenue}, \city{Nashville}, \postcode{37203}, \state{TN}, \country{USA}}}

\affil[2]{\orgdiv{Department of Biostatistics}, \orgname{Vanderbilt University Medical Center}, \orgaddress{\street{2525 West End Avenue}, \city{Nashville}, \postcode{37203}, \state{TN}, \country{USA}}}

\affil[3]{\orgdiv{Department of Pediatrics}, \orgname{Vanderbilt University Medical Center}, \orgaddress{\street{2200 Children's Way}, \city{Nashville}, \postcode{37232}, \state{TN}, \country{USA}}}

\affil[4]{\orgdiv{McWilliams School of Biomedical Informatics}, \orgname{The University of Texas Health Science Center at Houston}, \orgaddress{\street{7000 Fannin St \#600}, \city{Houston}, \postcode{77030}, \state{TX}, \country{USA}}}

\affil[5]{\orgdiv{Department of Medical Genetics and Genomics}, \orgname{Icahn School of Medicine at Mount Sinai}, \orgaddress{1 Gustave L. Levy Pl}, \city{New York}, \postcode{10028}, \state{New York}, \country{USA}}

\affil[6]{\orgdiv{Department of Medicine}, \orgname{Vanderbilt University Medical Center}, \orgaddress{\street{1161 21st Ave S}, \city{Nashville}, \postcode{37232}, \state{TN}, \country{USA}}}

\affil[7]{\orgdiv{Department of Computer Science}, \orgname{Vanderbilt University}, \orgaddress{\street{1400 18th Avenue S}, \city{Nashville}, \postcode{37212}, \state{TN}, \country{USA}}}

\affil[8]{\orgdiv{Department of Biomedical Informatics and Data Science}, \orgname{Yale School of Medicine}, \orgaddress{\street{101 College Street}, \city{New Haven}, \postcode{06520}, \state{CT}, \country{USA}}}

\equalcont{Equal contribution.}


\abstract{Phenotyping is fundamental to rare disease diagnosis, but manual curation of structured phenotypes from clinical notes is labor-intensive and difficult to scale. Existing artificial intelligence approaches typically optimize individual components of phenotyping but do not operationalize the full clinical workflow of extracting features from clinical text, standardizing them to Human Phenotype Ontology (HPO) terms, and prioritizing diagnostically informative HPO terms. We developed RARE-PHENIX, an end-to-end AI framework for rare disease phenotyping that integrates large language model-based phenotype extraction, ontology-grounded standardization to HPO terms, and supervised ranking of diagnostically informative phenotypes. We trained RARE-PHENIX using data from 2,671 patients across 11 Undiagnosed Diseases Network clinical sites, and externally validated it on 16,357 real-world clinical notes from Vanderbilt University Medical Center. Using clinician-curated HPO terms as the gold standard, RARE-PHENIX consistently outperformed a state-of-the-art deep learning baseline (PhenoBERT) across ontology-based similarity and precision-recall-F1 metrics in end-to-end evaluation (i.e., ontology-based similarity of 0.70 vs. 0.58). Ablation analyses demonstrated performance improvements with the addition of each module in RARE-PHENIX (extraction, standardization, and prioritization), supporting the value of modeling the full clinical phenotyping workflow. By modeling phenotyping as a clinically aligned workflow rather than a single extraction task, RARE-PHENIX provides structured, ranked phenotypes that are more concordant with clinician curation and has the potential to support human-in-the-loop rare disease diagnosis in real-world settings.}

\keywords{rare disease, artificial intelligence, large language model, phenotyping, clinical notes}

\maketitle

\section{Introduction}\label{sec1}

Rare diseases affect over 300 million individuals worldwide and are a major cause of chronic illness, disability, and premature mortality \citep{nguengang2020estimating}. While collectively common, rare diseases remain challenging to diagnose due to their low individual disease prevalence, limited clinician familiarity, and clinical heterogeneity \citep{nguengang2020estimating,valdez2016public,health2024landscape,schieppati2008rare}. As a result, many rare disease patients undergo prolonged diagnostic odysseys characterized by diagnostic delays averaging four to eight years, frequent misdiagnoses, and unnecessary testing \citep{evans2018dare,phillips2024time,faye2024time, tinker2024diagnostic}. Diagnostic odysseys have significant medical, psychosocial, and economic consequences for patients and families, resulting in irreversible disease progression, emotional turmoil, and avoidable healthcare expenditures \citep{morton2022importance, cohen2010quality,carmichael2015going,yang2022national}.

Phenotyping, the precise and systematic characterization of a patient’s clinical features, is fundamental to rare disease diagnosis and management. In practice, clinicians phenotype rare disease patients through a structured, multi-step clinical workflow: \textit{extract} clinical features from medical records, \textit{standardize} these features to Human Phenotype Ontology (HPO) terms, and \textit{prioritize} those that are diagnostically informative \citep{kohler2021human}. The resulting curated HPO terms serve as the foundation for diagnostic decision-making, including candidate gene and variant prioritization, iterative genomic analyses, and integration with additional clinical and functional evidence \citep{hartley2020new,wojcik2024genome,splinter2018effect,macnamara2020undiagnosed,philippakis2015matchmaker}. Though the widespread adoption of electronic health records (EHRs) provides new opportunities to support rare disease phenotyping through secondary data use, rare disease phenotypes are poorly captured in structured EHR fields because they are often atypical, heterogeneous, and inadequately represented by standard diagnostic or billing codes \citep{fung2014coverage,yadaw2025systematic}. Instead, they are often embedded in unstructured clinical notes, where manual curation is labor-intensive and difficult to scale.

Recent advances in artificial intelligence (AI) and large language models (LLMs) create new opportunities to support rare disease phenotyping at scale. Prior work has demonstrated the feasibility of these tools to extract rare disease phenotypes from unstructured clinical text, map clinical descriptions to ontologies, and support downstream diagnostic decision-making \citep{garcelon2018clinician, mak2024computer,dong2023ontology,segura2022exploring,xiao2023enhancing,zhao2026agentic,shyr2024identifying,shyr2025large,shyr2025accuracy,greco2026weakly,alsentzer2025few, yang2024enhancing,wu2025integrating, yang2025specialized,boceck2025aidiva,shyr2025statistical}. Early phenotyping approaches primarily relied on rule-based methods to identify rare disease phenotypes from EHRs \citep{garcelon2018clinician,mak2024computer,michalski2023supporting}. Subsequent work focused on supervised machine learning (ML) and deep learning models for phenotype extraction and normalization \citep{segura2022exploring,dong2023ontology,lievin2023findzebra}. More recent studies demonstrated that LLMs, particularly when augmented with domain knowledge, ontologies, or retrieval mechanisms, can generate phenotypic summaries, rank candidate diseases or genes, or assist with rare disease diagnosis \citep{mao2025phenotype,yang2025rdguru,cao2024automatic,wen2025phenodp,kim2024assessing,young2025diagnostic,shyr2025large,shyr2025accuracy,yang2025rdguru}. Benchmarking studies further suggest that LLMs can achieve comparable or exceed clinician-level performance on rare disease reasoning tasks when provided with carefully designed prompts or external knowledge sources \citep{chen2025enhancing,zhong2025performance,rider2025evaluating}.

However, most existing approaches optimize \textit{individual} components of rare disease phenotyping in isolation rather than supporting the \textit{end-to-end clinical workflow}. In clinical practice, the challenge is not only to extract phenotypes, but also prioritize the subset that meaningfully narrows the differential diagnosis. Rare disease patients often present with common or non-specific phenotypes (e.g., fatigue) that are less diagnostically informative. Therefore, a phenotyping tool may have high accuracy but remain poorly suited for diagnosis because informative phenotypes are diluted by a long list of true but diagnostically non-related terms. Many current methods focus on phenotype extraction or normalization without distinguishing which phenotypes are most relevant for diagnosis \citep{segura2022exploring,shyr2024identifying,xiao2023enhancing,dong2023ontology, garcia2025improving, wu2024hybrid, liu2022oard, groza2024evaluation, thompson2023large}. Others perform disease or gene prioritization using phenotypes as input, focusing on downstream interpretation rather than phenotype extraction from unstructured clinical notes \citep{kim2024assessing}. Recent LLM-based diagnostic systems and retrieval-augmented frameworks primarily focus on diagnostic decision support, rather than modeling the end-to-end phenotyping workflow performed by clinicians \citep{zhong2025performance,wen2025phenodp,chen2025enhancing, zhao2026agentic, mao2025phenotype}. As a result, to our knowledge, no existing approach provides an end-to-end rare disease phenotyping pipeline aligned with real-world clinical practice.

To address this gap, we developed and externally validated \textbf{RARE-PHENIX} (\textbf{RARE} disease \textbf{PHEN}otyping with \textbf{I}ntelligent e\textbf{X}traction), a modular AI framework designed to perform end-to-end rare disease phenotyping from unstructured clinical notes to support diagnosis. RARE-PHENIX aligns with real-world clinical workflows by 1) \textit{extracting} rare disease phenotypes using LLMs with few-shot prompting and instruction fine tuning; 2) \textit{standardizing} them to HPO terms using retrieval-augmented generation; and 3) \textit{prioritizing} HPO terms using a supervised ranking model trained to distinguish common, non-specific phenotypes from those that are diagnostically informative. We trained RARE-PHENIX on rare disease-specific corpora, including expert-annotated documents from the National Organization of Rare Disorders database and synthetic clinical text containing clinician-curated HPO terms across 11 Undiagnosed Diseases Network (UDN) clinical sites. We then externally validated RARE-PHENIX on 16,357 real-world clinical notes from patients at Vanderbilt University Medical Center (VUMC), which were not included among the training sites. Across evaluations, RARE-PHENIX outperformed PhenoBERT \citep{feng2022phenobert}, a state-of-the-art system for identifying HPO terms from unstructured text, at extracting diagnostically relevant rare disease phenotypes. By aligning AI-based phenotyping with real-world clinical workflows, RARE-PHENIX enables end-to-end rare disease phenotyping from clinical notes to support diagnosis. 

\section{Methods}
\subsection{Study Design and Population} This is a retrospective, multi-site cohort study with model development and external validation. The study included $N = 2,814$ patients evaluated at the UDN between October 4, 2010 and April 25, 2024 across 12 clinical sites: University of California, Los Angeles, Baylor College of Medicine, Duke University, Stanford University, Harvard-affiliated Hospitals, NIH Undiagnosed Diseases Program, Washington University in St. Louis, Children’s Hospital of Philadelphia and University of Pennsylvania, University of Washington and Seattle Children’s Hospital, University of Utah, and University of Miami, and VUMC. In general, patients are accepted to the UDN if they have rare or undiagnosed diseases resulting in multisystem dysfunction, functional impairment, or symptom severity that substantially affects quality of life. Eligibility requires objective or measurable findings from previous testing and examinations, and the absence of a diagnosis after prior workup.

For model development, we used data from 11 UDN sites ($N = 2,671$). The data from VUMC ($N = 143$) were held out as an external test cohort and was not used at any stage of model training or selection. We restricted our analysis to patients who have at least one UDN clinician-curated HPO term. All data partitioning was performed at the patient level to prevent data leakage. This study was approved by the central UDN Institutional Review Board (IRB \#172005) and VUMC Institutional Review Board (IRB \#222249). All protected health information was processed within secure, institutionally approved environments in accordance with data governance policies. Study reporting followed the TRIPOD-AI checklist \citep{collins2024tripod+}.

\subsection{Overview of RARE-PHENIX} 
\begin{figure}[!htbp]
    \centering
       \caption{\textbf{Overview of RARE-PHENIX}. RARE-PHENIX is an end-to-end AI system for automating the extraction, standardization, and prioritization of rare disease phenotypes from unstructured clinical text. This system consists of three modules for 1) extracting rare disease features from clinical notes with large language models (LLMs); 2) standardizing these features to structured Human Ontology Phenotype (HPO) terms using retrieval-augmented generation; and 3) prioritize diagnostically informative HPO terms using a supervised ranking model. LLMs include \texttt{LLaMA-2-chat (7b, 13b, and 70b)}, \texttt{LLaMA-3-instruct (8b, and 70b) LLaMA-3.1-instruct (8b and 70b), LLaMA-3.2-instruct (1b and 3b), LLaMA-3.3-instruct (70b)}, and a secure instance of Azure OpenAI's \texttt{ChatGPT-4o} (v2024-06-01) provisioned for handling protected health information in accordance with institutional data governance policies.}
    \includegraphics[width=1\linewidth]{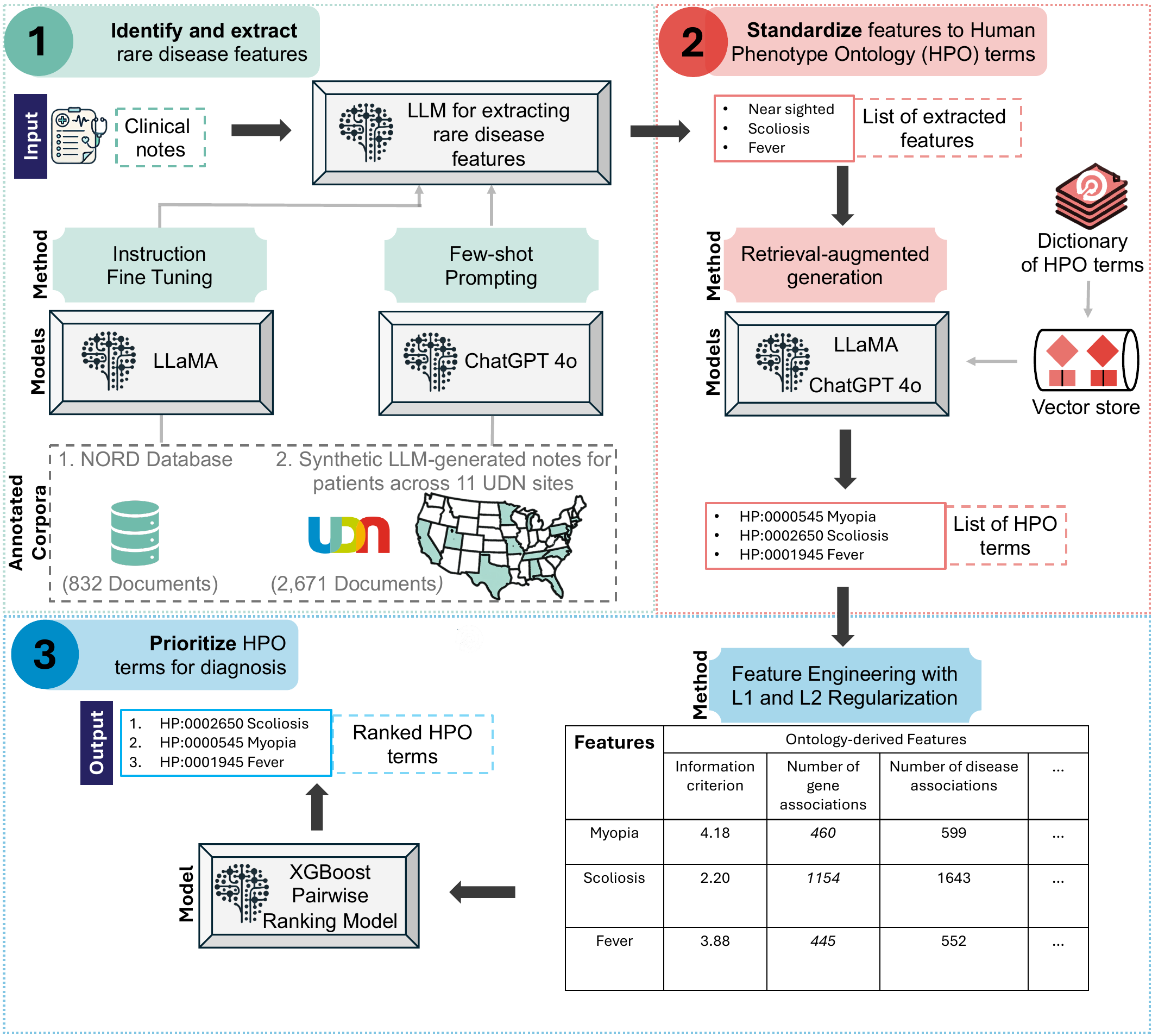}
    \label{fig:1}
\end{figure}

RARE-PHENIX is an end-to-end AI framework that automates phenotype extraction from clinical notes, standardizes extracted features to HPO terms, and prioritizes diagnostically informative HPO terms. The framework consists of three sequential modules (\textbf{Fig.~\ref{fig:1}}). 

\subsection{Module 1: Phenotype Extraction from Clinical Notes} The objective of Module 1 is to identify and extract rare disease phenotypes from unstructured clinical notes. We developed two complementary approaches: parameter-efficient instruction fine-tuning of open-source LLMs and few-shot prompting of a closed-source model, enabling evaluation under both deployable and API-based settings.

\subsubsection{Instruction Fine-Tuning} Instruction fine-tuning adapts a pre-trained LLM to a domain-specific task using labeled instruction–output pairs \citep{zhang2026instruction}. We fine-tuned 10 LLaMA models of various sizes: \texttt{LLaMA-2-chat (7b, 13b, and 70b), LLaMA-3-instruct (8b, and 70b) LLaMA-3.1-instruct (8b and 70b), LLaMA-3.2-instruct (1b and 3b)}, and \texttt{LLaMA-3.3-instruct (70b)} \citep{touvron2023llama}. We selected LLaMA for its strong performance on biomedical and clinical natural language processing (NLP) tasks, open-source availability, and compatibility with parameter efficient fine tuning (PEFT) \citep{xie2025medical, hu2026information}. LLaMA models were fine-tuned using PEFT. Each training instance consisted of an input text and a corresponding HTML-annotated version in which rare disease phenotypes were wrapped in \texttt{<span>} tags. Instruction fine-tuning was performed using the HuggingFace Transformers and TRL framework with parameter-efficient fine-tuning (PEFT) \citep{touvron2023llama}. Base models were fine-tuned using QLoRA, with 4-bit NF4 weight quantization and bfloat16 computation. Low-Rank Adaptation (LoRA) was applied with rank $r=16$, scaling factor 64, and dropout 0.05 to the linear projection layers, reducing computational requirements \citep{dettmers2023qlora}. Training was conducted using TRL’s \texttt{SFTTrainer} with a maximum sequence length of 1{,}500 tokens for two epochs, a learning rate of $2\times10^{-4}$, and a warm-up ratio of 0.05. Fine-tuning was performed using 3x80GB A100 NVIDIA GPUs.

\textit{Fine-Tuning Corpora.} We used two complementary corpora:

\begin{enumerate}
    \item \textbf{RareDis Corpus} \citep{martinez2022raredis}. This corpus consists of 832 expert-annotated documents from the National Organization for Rare Disorders knowledgebase. Documents were annotated by experts for four entity types (i.e., rare disease, disease, symptom, and sign) with an inter-annotator agreement F1-score of 83.5\%, indicating high annotation reliability. RareDis documents were converted into instruction-output pairs by wrapping the annotated rare disease entities in HTML span tags. 
    \item \textbf{UDN Synthetic Clinical Narratives.} To augment limited annotated data with clinically grounded narratives, we generated synthetic clinical text using clinician-curated HPO terms from the $N=2,671$ patients in the training cohort. We generated synthetic clinical text using a secure, institutionally approved instance of Azure OpenAI’s \texttt{ChatGPT-4 turbo} (v2024-04-09) with a temperature of zero for reproducibility. Each patient's curated HPO term was incorporated verbatim in the synthetic clinical text and wrapped in HTML tags. The full prompt used to generate these documents is provided in the \textbf{Supplementary Materials}.  
\end{enumerate}

\subsubsection{Few-Shot Prompting Approach} Few-shot prompting was performed using a secure instance of Azure OpenAI’s \texttt{ChatGPT-4o} (v2024-06-01) with temperature set to 0. Prompts included a task statement, markup guide, detailed definition of rare disease phenotypes, a set of labeled examples, and the input text. The model was instructed to output an HTML-annotated version of the text, with rare disease phenotypes wrapped in \texttt{<span>} tags. We provide the full prompt in \textbf{Supplementary Table S1}.

\subsection{Module 2: Standardization to HPO Terms} The objective of Module 2 is to standardize phenotype strings to structured HPO terms (i.e., ``near sighted" $\rightarrow$ HP:0000545 Myopia) using retrieval-augmented generation (RAG) \citep{lewis2020retrieval}. This module produces standardized, interoperable phenotype terms that can be used for downstream diagnostic workup (e.g., genomic analysis).

RAG combines semantic retrieval with LLM-based generation to ground model outputs in external knowledge such as the HPO. We chose RAG to address limitations of both retrieval-only methods, which may miss contextual nuance in clinical descriptions, and unconstrained LLMs, which may hallucinate invalid HPO terms. We constructed a vector database of HPO terms, where each term was represented as a text node and embedded using a pre-trained sentence embedding model (i.e., \texttt{text-embedding-3-small}). The embeddings were indexed in a vector store using cosine similarity. For each phenotype string, Module 2 performs the following:
\begin{enumerate}
    \item \textit{Semantic retrieval:} The phenotype string is embedded and used to retrieve the top-10 semantically similar HPO terms from the vector store based on cosine similarity.
    \item \textit{Ontology-grounded generation:} The retrieved HPO nodes are provided to an LLM, which is prompted to select the most appropriate HPO term based on contextual relevance and the associated ontology definition. If no suitable match is found, the model outputs ``none."
\end{enumerate}

\subsection{Module 3: Prioritization of Diagnostically Informative Phenotypes} The objective of \textbf{Module 3} is to operationalize the prioritization of diagnostically informative phenotypes as a supervised learning-to-rank task. Phenotypes are defined as informative if they were uncommon, clinically specific, and associated with a limited number of diseases or genes, making them more effective for distinguishing among candidate diagnoses. 

\subsubsection{Learning-to-Rank Formulation} We formulated HPO term prioritization as a supervised learning-to-rank problem in which, for each patient, the model learns to assign higher relevance scores to clinician-curated HPO terms than non-clinician-curated terms. The training cohort included $N=2,671$ patients from 11 UDN clinical sites, excluding VUMC. For each patient, clinician-curated HPO terms were labeled as positives (label = 1). Because these HPO terms are not provided in a ranked order, we trained the learning-to-rank model to rank them above non-clinician-curated terms, or negative HPO terms, within each patient. We constructed negative HPO terms (label = 0) for each patient, featuring a mix of difficult (very similar to the clinician-curated HPO terms), medium (somewhat similar), easy (not very similar), and implausible negatives (highly unrelated). Difficult negatives were randomly sampled from sibling and cousin terms in the ontology, medium negatives from terms three to five edges away, easy negatives from distant ancestor or descendant terms, and implausible negatives from terms sharing no close ancestry ($\leq 2$ edges) with any positive term.

\textit{Feature Engineering.} Model features included patient-level predictors (i.e., age, sex, primary symptom category) and knowledgebase-derived annotations from Online Mendelian Inheritance in Man (OMIM) and Orphanet \citep{amberger2019omim}. For each HPO term, features included information content, numbers and fractions of associated genes and diseases, and OMIM- and Orphanet-based inverse document frequencies (IDF). IDFs were calculated as the negative logarithm of the fraction of associated diseases. Collectively, these knowledgebase-derived annotations capture phenotype rarity and specificity, with higher values indicating greater diagnostic informativeness.

\textit{Model Training and Selection} We randomly partitioned the training data at the patient level into an 80:20 training and validation set. Using the training set, we trained multiple supervised models, including gradient-boosted decision tree learning-to-rank approaches (XGBoost with a pairwise ranking objective, LightGBM with a LambdaRank, and CatBoost with YetiRankPairwise) as well as a logistic regression ranker baseline \citep{chen2016xgboost,ke2017lightgbm,prokhorenkova2018catboost}. Hyperparameters for the gradient boosted models, including learning rate, tree depth, and $\ell_1$, $\ell_2$ regularization, were tuned using early stopping based on mean average precision (MAP) on the validation set. Model performance was also validated using MAP at 30 (MAP@30), computed at the patient level. The same validation split and metric were applied uniformly across all models to ensure fair comparison. The model with the highest validation MAP@30 was selected as the final ranking model and applied to the held-out, external test cohort for evaluation. 

\subsection{Evaluation Strategy and Baseline Comparator} \textit{Evaluation Metrics.} We evaluated RARE-PHENIX on an external cohort of 143 UDN patients at VUMC, who had a total of 16,357 clinical notes recorded prior to the UDN workup. The primary outcome was phenotypic concordance with clinician-curated HPO terms, measured using ontology-based semantic similarity (Lin measure) \citep{lin1998information}. Secondary outcomes included term-level phenotyping performance as measured by precision, recall, and F1 score.

\textit{Clinical Notes Pre-processing.} To prevent data leakage, we excluded clinical notes recorded after the start of UDN’s evaluation. Notes were cleaned to remove administrative content (e.g., scheduling notes) and segmented into chunks of at most 4,026 characters while respecting sentence boundaries. This chunk size was chosen to accommodate the smallest context window among the LLMs used.

\textit{Baseline Comparator.} We compared RARE-PHENIX to PhenoBERT, a state-of-the-art deep learning system for HPO term extraction \citep{feng2022phenobert}. PhenoBERT uses a two-stage architecture: 1) a hierarchical convolutional neural network for candidate HPO term selection, followed by 2) a BERT-based model for candidate evaluation. In published benchmarks, PhenoBERT outperformed multiple dictionary-based systems and prior deep learning methods, including NeuralCR and PhenoTagger, and achieved state-of-the-art performance across both PubMed abstracts and real-world clinical notes, making it an ideal baseline comparator \citep{luo2021phenotagger, arbabi2019identifying}.

\section{Results}
This study included $N=2{,}814$ patients evaluated at the UDN. RARE-PHENIX was developed using data from 11 clinical sites ($N=2{,}671$, 94.9\%) and externally validated on an independent cohort from VUMC ($N=143$, 5.1\%), which was not used at any stage of model training or selection. The cohort was primarily pediatric, with a median age of 12 years (IQR: 5–31), and had a median of 15 clinician-curated HPO terms per patient (IQR: 9–25), reflecting substantial phenotypic complexity (\textbf{Table~\ref{tab:1}}).  

External validation was performed on 16,357 clinical notes, the majority of which were progress notes (68\%), followed by consultation notes (13\%) and procedure or operative notes (6\%) (\textbf{Table~\ref{tab:2}}). Patients had a median of 18 notes each (IQR: 9–113.5), with a median note length of 1,792 characters (IQR: 573–4,297).
\begin{table*}[h]
\centering
\eightpt
\caption{\textbf{Overview of Undiagnosed Diseases Network (UDN) data}. HPO = Human Phenotype Ontology. IQR = interquartile range.}
\label{tab:1}

\begin{tabularx}{\textwidth}{lYYY}
\toprule
\textbf{Clinical Site} & \textbf{Number of patients, n (\%)} & \textbf{Age, median (IQR)} & \textbf{Number of HPO terms, median (IQR)} \\
\midrule

\multicolumn{4}{l}{\textbf{Training Cohorts}} \\

UCLA & 248 (8.8\%) & 11.0 (4.2--23.8) & 22.0 (13.0--38.2) \\
Baylor & 358 (12.7\%) & 8.0 (3.0--18.0) & 16.0 (9.0--25.0) \\
Duke & 302 (10.7\%) & 7.0 (3.0--14.5) & 21.5 (12.0--31.0) \\
Stanford & 325 (11.5\%) & 11.0 (5.0--29.8) & 10.0 (6.0--17.0) \\
Harvard-affiliate & 240 (8.5\%) & 12.5 (4.0--29.0) & 13.0 (7.0--22.0) \\
NIH & 636 (22.6\%) & 25.0 (10.0--46.0) & 21.0 (12.0--35.0) \\
WUSTL & 122 (4.3\%) & 10.0 (3.0--20.5) & 12.0 (8.2--17.0) \\
CHOP-UPenn & 134 (4.8\%) & 14.0 (4.0--35.0) & 8.0 (5.2--11.0) \\
UW-SCH & 154 (5.5\%) & 11.0 (4.0--34.0) & 11.0 (7.0--17.0) \\
Utah & 65 (2.3\%) & 16.0 (4.0--37.0) & 7.0 (5.0--11.0) \\
Miami & 87 (3.1\%) & 11.0 (4.0--20.0) & 14.0 (9.0--20.0) \\

\midrule
\multicolumn{4}{l}{\textbf{Evaluation Cohort}} \\

Vanderbilt & 143 (5.1\%) & 8.0 (3.0--20.5) & 17.0 (12.0--23.0) \\

\midrule
\textbf{Overall} & 2814 (100.0\%) & 12.0 (5.0--31.0) & 15.0 (9.0--25.0) \\

\bottomrule
\end{tabularx}
\end{table*}

\begin{table*}[h]
\centering
\eightpt
\caption{\textbf{Summary of clinical notes in the external evaluation cohort}. IQR = interquartile range.}
\label{tab:2}

\begin{tabularx}{\textwidth}{
l
>{\raggedleft\arraybackslash}X
>{\raggedleft\arraybackslash}X
>{\raggedleft\arraybackslash}X}
\toprule
\textbf{Clinical note type} & \textbf{N (\%)} & \textbf{Number of notes per patient, median (IQR)} & \textbf{Number of characters per note, median (IQR)} \\
\midrule

Progress & 11083 (67.8\%) & 16.0 (8.0--71.5) & 2326.0 (819.5--5008.0) \\
Consultations & 2074 (12.7\%) & 9.0 (2.5--27.0) & 901.0 (506.2--2200.5) \\
Procedures / Operative & 926 (5.7\%) & 9.5 (4.0--17.8) & 942.0 (205.0--2189.2) \\
Imaging & 712 (4.4\%) & 4.0 (2.0--14.0) & 2276.5 (569.5--3376.2) \\
Rehabilitation \& Therapy & 298 (1.8\%) & 4.0 (1.2--12.0) & 2993.5 (610.0--4137.8) \\
Nutrition & 169 (1.0\%) & 2.0 (1.0--14.0) & 349.0 (221.0--588.0) \\
Letters \& Correspondence & 166 (1.0\%) & 2.0 (1.0--2.5) & 1710.5 (354.0--3395.8) \\
Diagnostics (Non-Imaging) & 98 (0.6\%) & 6.0 (3.0--13.0) & 1051.5 (301.0--2083.0) \\
Discharge & 54 (0.3\%) & 1.5 (1.0--3.0) & 4468.0 (2512.2--6008.5) \\
Other & 777 (4.8\%) & 6.5 (3.0--22.5) & 406.0 (181.0--707.0) \\

\midrule
\textbf{Combined} & \textbf{16357 (100.0\%)} & \textbf{18.0 (9.0--113.5)} & \textbf{1792.0 (573.0--4297.0)} \\

\bottomrule
\end{tabularx}
\end{table*}

\subsection{End-to-End Rare Disease Phenotyping Performance} 
We evaluated the end-to-end performance of RARE-PHENIX (phenotype extraction, HPO standardization, and prioritization) on the external validation cohort and compared it to PhenoBERT using clinician-curated HPO terms as the ground truth. Because PhenoBERT does not rank HPO terms by design, we passed PhenoBERT-extracted HPO terms through the same prioritization module (Module 3) to generate top-$k$ lists for a head-to-head comparison. Within Module 3, XGBoost achieved the highest validation performance (MAP@30 = 0.85) and was selected as the final learning-to-rank model for external evaluation (\textbf{Supplementary Table S3}).

\begin{figure}[!htbp]
    \centering
      \caption{\textbf{End-to-end performance results of RARE-PHENIX and PhenoBERT on the external validation cohort.} For legibility, only the top-performing large language models are shown in the figure (i.e., \texttt{ChatGPT-4o, LLaMA-2-70b, LLaMA-3-70b, LLaMA 3.1-70b}) in addition to the baseline comparator (PhenoBERT) across top-$k$ cutoffs. The end-to-end performance results of other RARE-PHENIX configurations are provided in \textbf{Supplementary Table S2}.}
    \includegraphics[width=1\linewidth]{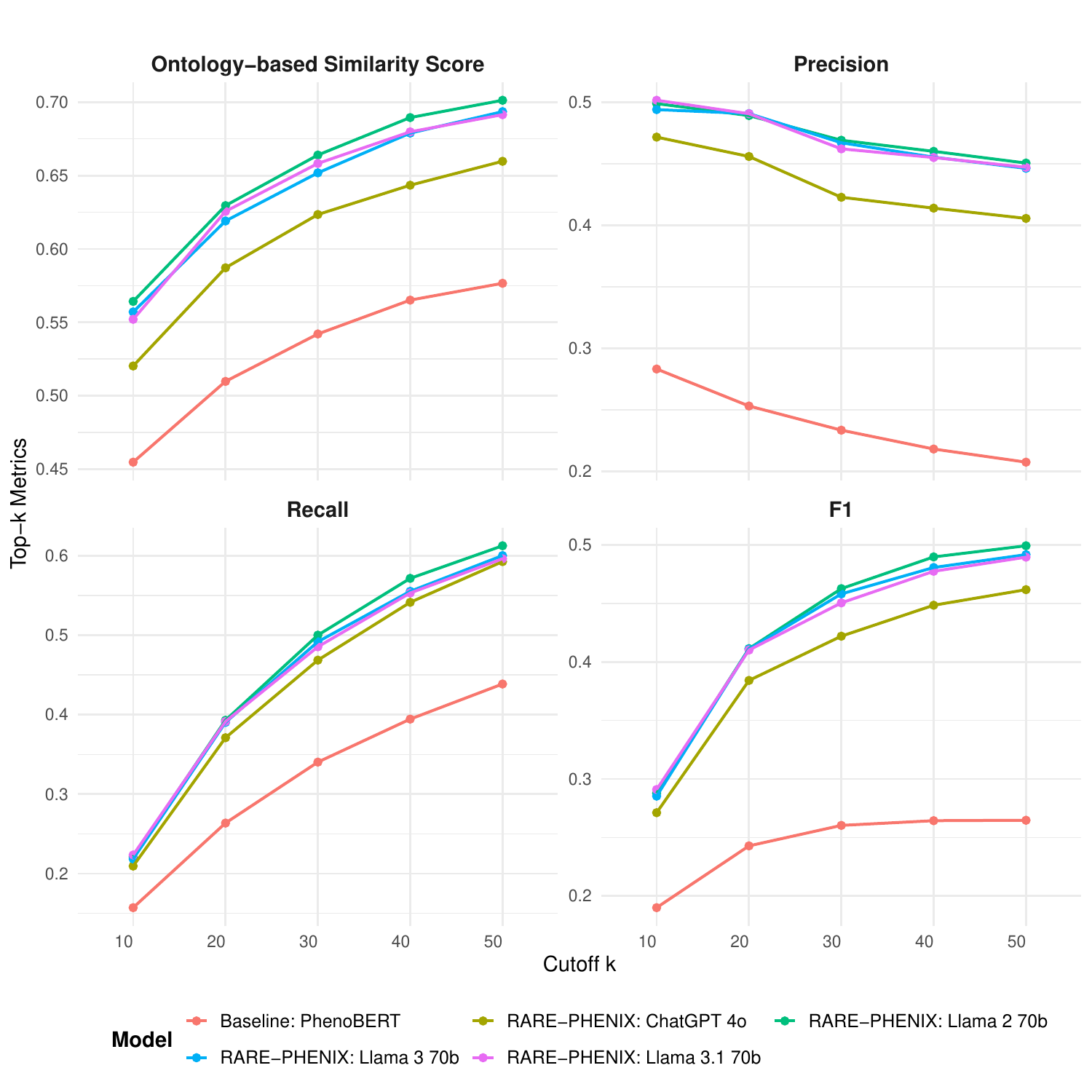}
    \label{fig:2}
\end{figure}

\textbf{Fig.~\ref{fig:2}} summarizes the end-to-end performance across top-$k$ cutoffs ($k = 10, 20, \ldots, 50$); for legibility, the figure displays only the top-performing LLM configurations, with full results provided in \textbf{Supplementary Table S2}. Across all cutoffs and metrics, RARE-PHENIX consistently outperformed the PhenoBERT baseline. In particular, RARE-PHENIX achieved higher ontology-based semantic similarity to clinician-curated phenotypes, with the strongest similarity observed for \texttt{LLaMA-2-70b} across cutoffs (e.g., at $k=50$, $\sim$0.70 for \texttt{LLaMA-2-70b} versus $\sim$0.58 for PhenoBERT), indicating improved concordance with clinician-curated phenotypes. 

Consistent with expected retrieval trade-offs, precision decreased as $k$ increased while recall increased, reflecting that broader phenotype lists contained more clinician-curated terms at the cost of including additional non-curated terms. Notably, F1 improved sharply from $k=10$ to $k=20$ and then plateaued by approximately $k=30$--$40$, suggesting diminishing benefit from including additional terms beyond this range for balanced performance. Among the top-performing configurations, \texttt{LLaMA-2-70b} generally achieved the highest recall, F1, and ontology-based similarity across all cut-offs.

\FloatBarrier

\subsection{Module-based Ablation Analysis of RARE-PHENIX} 
\begin{figure}[!htbp]
    \centering
       \caption{\textbf{Module-based Ablation Analysis Results of RARE-PHENIX Across Extraction and Standardization Modules}}
    \includegraphics[width=1\linewidth]{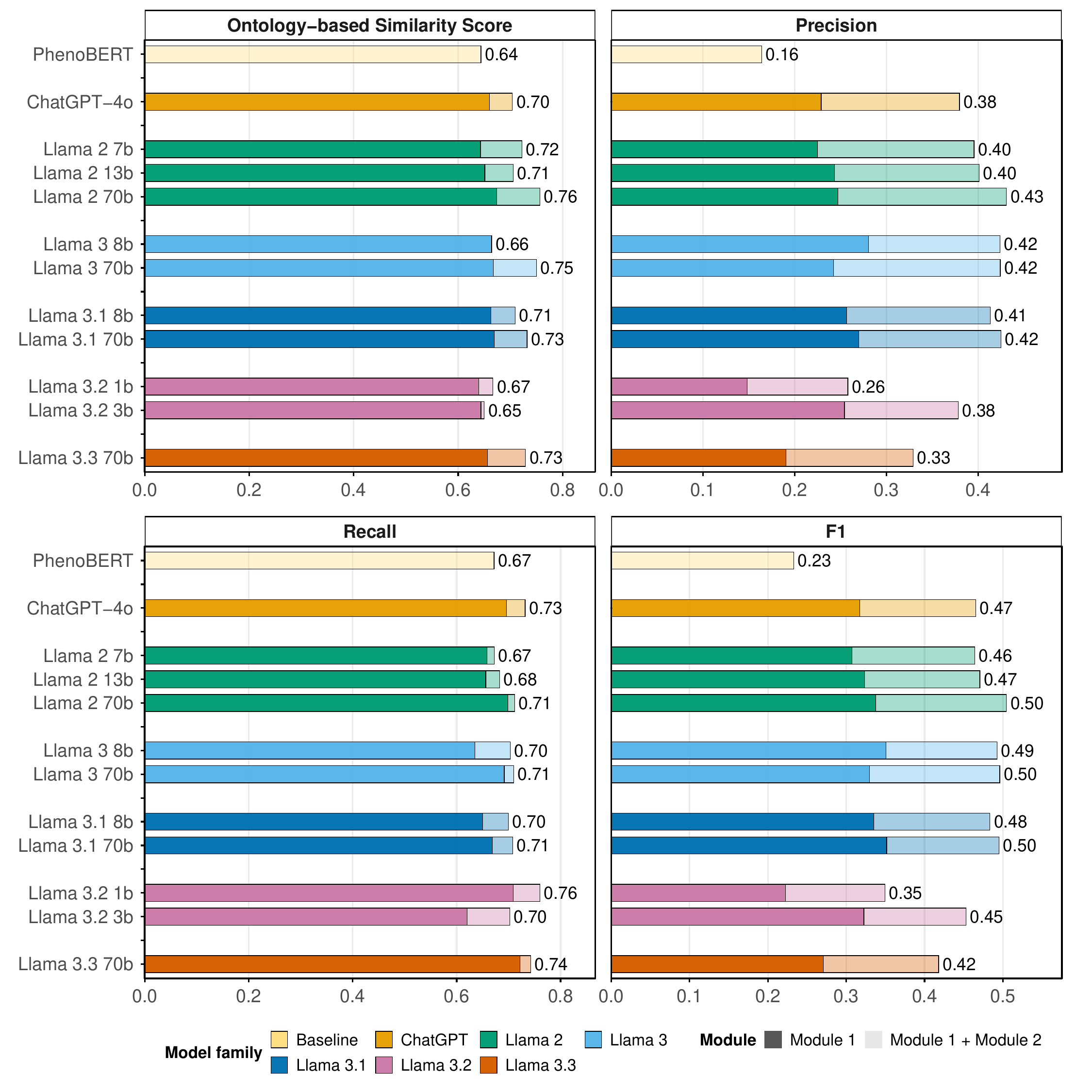}
    \label{fig:3}
\end{figure}

\textbf{Fig.~\ref{fig:3}} illustrates the module-based ablation analysis results for extraction alone (Module 1) and extraction followed by HPO standardization (Modules 1+2), without prioritization. Across model configurations, standardization (Module 2) consistently improved phenotype concordance with clinician-curated lists, with the largest gains observed in precision and F1, indicating that mapping free-text phenotypes to HPO terms substantially reduces noise and yields more specific phenotype representations.

Among all RARE-PHENIX configurations, \texttt{LLaMA-2-70b} achieved the highest ontology-based similarity (0.76), followed by \texttt{LLaMA-3-70b} (0.75) and both \texttt{LLaMA-3.1-70b} and \texttt{LLaMA-3.3-70b} (0.73), compared to 0.70 for few-shot \texttt{ChatGPT-4o} and 0.64 for PhenoBERT. In general, PhenoBERT performed similarly to smaller LLMs (1-3B parameters), whereas larger models (70b) consistently achieved higher ontology-based similarity.

Standardization yielded substantial improvements in precision across most LLMs (e.g., \texttt{LLaMA-2-70b} increased from 0.25 with Module 1 to 0.43 with Modules 1+2), translating into marked gains in F1 (from 0.34 to 0.50). In contrast, recall was relatively stable, with smaller gains after standardization (e.g., \texttt{LLaMA-2-70b} 0.70 to 0.71), consistent with Module 2 primarily improving specificity rather than simply increasing coverage. Notably, some smaller models achieved high recall with lower precision (e.g., \texttt{LLaMA-3.2-1b} recall 0.76 with precision 0.26), illustrating how phenotype lists can recover many clinician-curated terms while remaining diluted by less informative or extraneous phenotypes. Together, these results identify ontology-based standardization as a key contributor to improved phenotype concordance.

\FloatBarrier
\subsection{Contribution of Phenotype Prioritization Module}
To assess the isolated contribution of the phenotype prioritization module (Module~3), we compared the ranked phenotype lists produced by RARE-PHENIX to a random ordering of the same extracted phenotypes. For each patient, the extracted HPO terms were randomly permuted 200 times, and performance was evaluated at top-$k$ cutoffs ($k = 10, 20, 30, 40, 50$). Metrics were computed at the patient level with 95\% bootstrap intervals across 1{,}000 bootstrap iterations. Improvements attributable to prioritization were quantified as the difference between prioritized and randomly ordered lists ($\Delta$ = prioritized using Module 3 $-$ prioritized by random ordering).

\begin{figure}[!htbp]
    \centering
        \caption{\textbf{Contribution of phenotype prioritization to diagnostic utility.}
Improvement in performance using the prioritization module (Module 3) relative to a random ordering of the same extracted phenotypes. For each patient, extracted HPO terms were randomly permuted 200 times, and performance was evaluated at top-$k$ cutoffs ($k = 10, 20, 30, 40, 50$). Values represent the mean difference ($\Delta$ = ranking by Module 3 $-$ ranking by random ordering), and shaded regions indicate 95\% bootstrap intervals obtained by resampling at the patient level.}
    \includegraphics[width=1\linewidth]{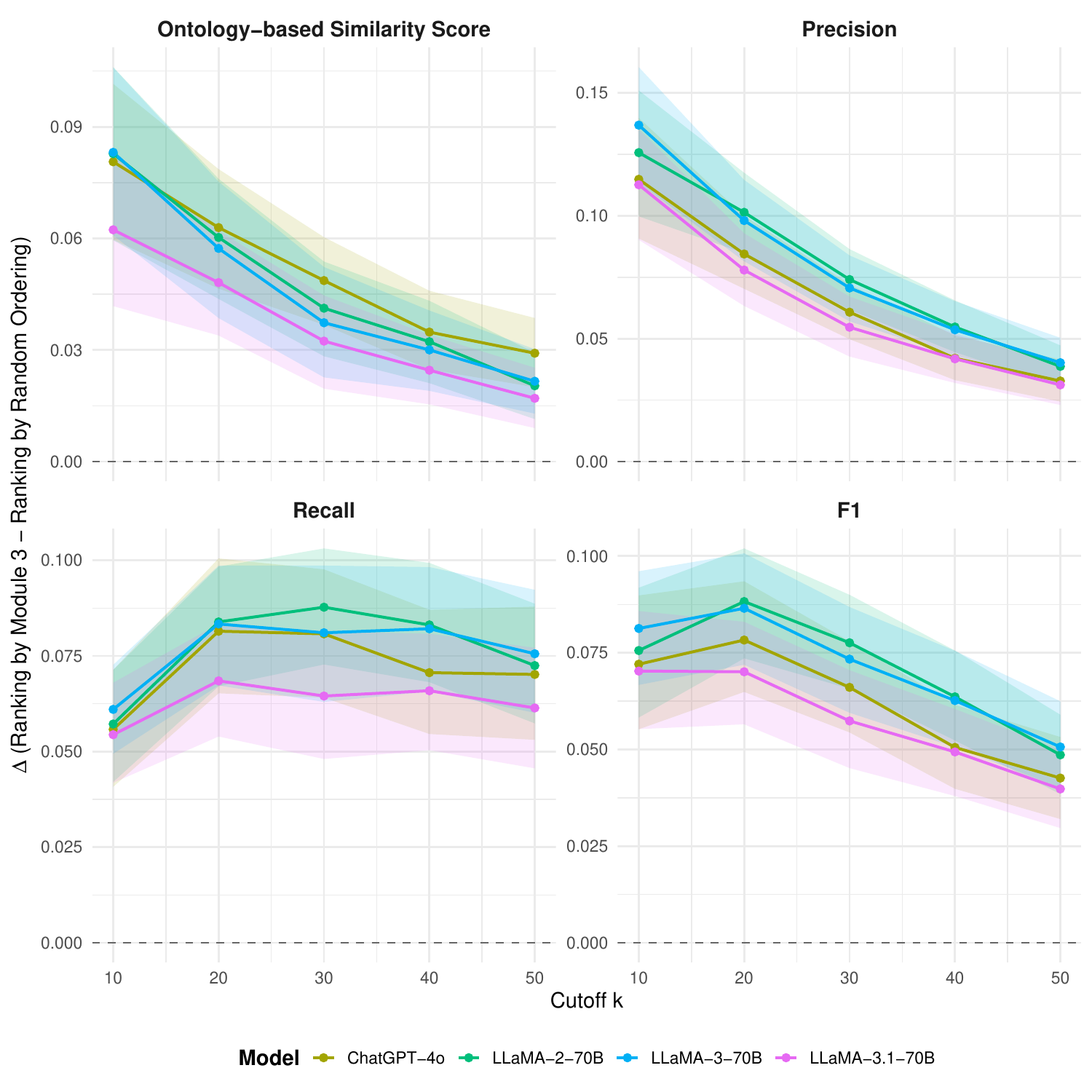}
    \label{fig:4}
\end{figure}

Our results show that phenotype prioritization (Module 3) consistently improved performance across all models and cutoffs (\textbf{Fig.~\ref{fig:4}}). The largest gains were observed at lower $k$, where clinical decision-making typically focuses on a limited number of highly informative phenotypes. At $k = 10$, prioritization increased ontology-based similarity by approximately 0.06-0.09 and precision by 0.11-0.14, indicating that the highest-ranked phenotypes were substantially more diagnostically relevant than would be expected by chance. Improvements in recall and F1 score were also observed, with highest gains at intermediate cutoffs ($k$ = 20–30). We provide the results with 95\% bootstrap intervals for all models in \textbf{Supplementary Table S4}.

The magnitude of the prioritization benefit decreased as $k$ increased, which was expected because the difference between prioritized and random ordering diminishes when most extracted phenotypes are included. Importantly, the lower bound of the 95\% bootstrap intervals remained above zero across all cutoffs and models, demonstrating that the observed gains were consistent at the patient level. Collectively, these results show that Module~3 systematically prioritizes diagnostically informative phenotypes that align with real-world clinician curation.

\subsection{Systematic Error Analysis}
\begin{figure}[!htbp]
    \centering
       \caption{\textbf{Results of systematic error analysis}. False negatives and false positives of RARE-PHENIX with the best-performing large language model configurations (\texttt{ChatGPT-4o, LLaMA-2-70b, LLaMA-3-70b, LLaMA-3.1-70b}) and the baseline comparator (PhenoBERT) at different top-k cutoffs $(k = 10, 20, 30, 40, 50)$.}
    \includegraphics[width=1\linewidth]{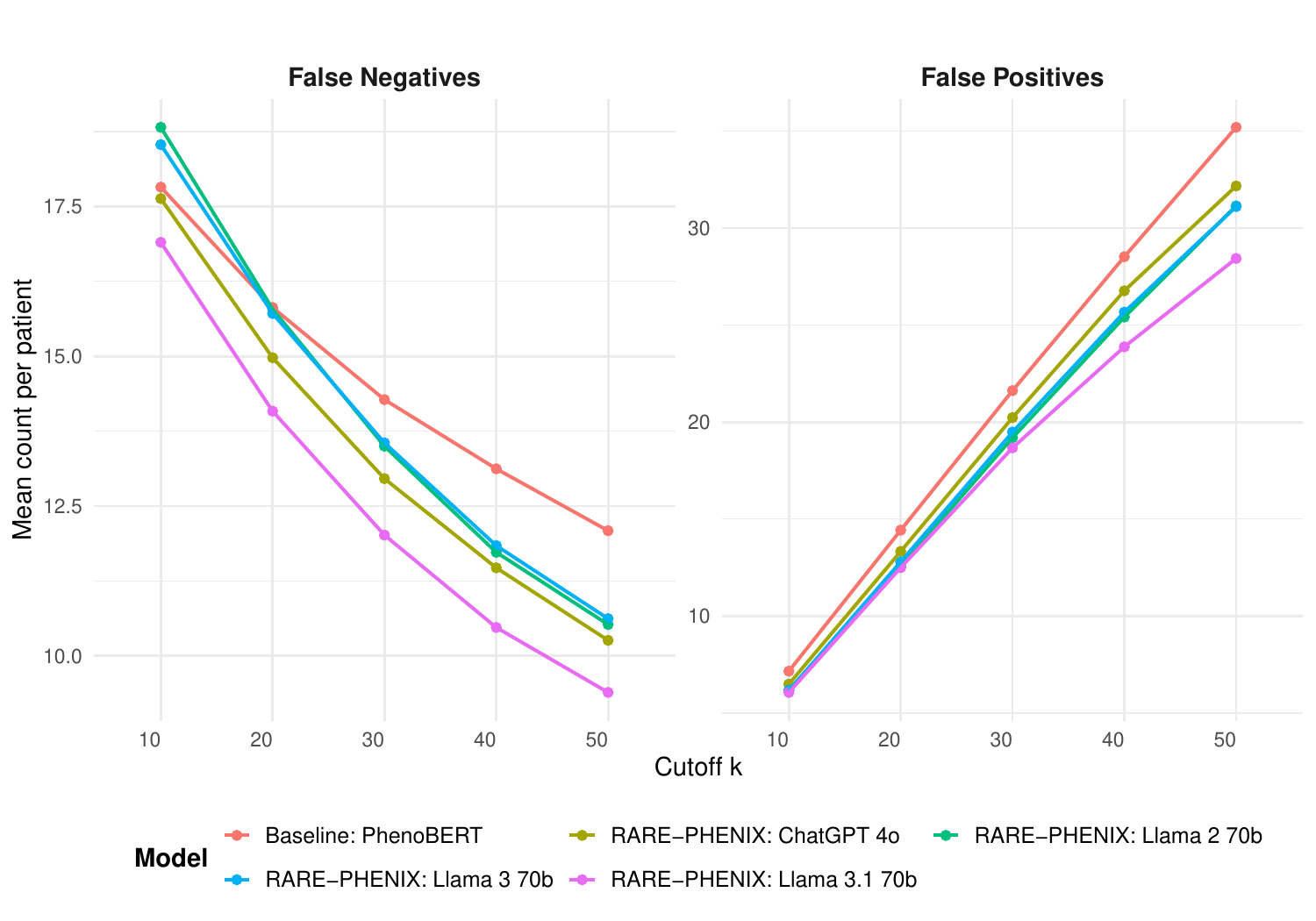}
    \label{fig:5}
\end{figure}

To characterize the systematic errors of RARE-PHENIX, we conducted a clinician-adjudicated systematic error analysis of false negatives (FNs) and false positives (FPs) across top-$k$ cutoffs. RARE-PHENIX consistently produced fewer errors than PhenoBERT, with a 29\% relative reduction in mean FNs per patient and a 24\% reduction in mean FPs per patient at $k=50$ (\textbf{Fig.~\ref{fig:5}}).

Manual review of 271 FN instances revealed that the majority (97\%) were not due to extraction failures, but rather to the way phenotypes are documented in real-world clinical narratives. Most missed terms were expressed indirectly through linguistic variation or contextual description. For example, the HPO term \textit{failure to thrive} was described in clinical notes as ``severe growth retardation,'' and \textit{short foot} as ``both feet appear to be small.'' In several cases, the model extracted a semantically related or higher-level HPO concept. Additional FNs reflected limitations of clinical documentation, such as physical examination findings that were not explicitly recorded in the notes. Only 3\% of FNs were attributable to true extraction errors in which clearly documented HPO terms were not identified by the model. FPs increased with larger $k$ values for all methods, reflecting the expected trade-off between coverage and specificity. Manual review revealed that FPs primarily occurred due to ontology granularity differences (i.e., extracting the parent term rather than the child term) and common, non-specific symptoms that were not included in clinician-curated lists (e.g., ``nausea"). RARE-PHENIX maintained consistently lower FP counts than PhenoBERT across all cutoffs, indicating more precise phenotyping (\textbf{Fig.~\ref{fig:5}}). Collectively, these findings suggest that errors are driven primarily by the linguistic and structural characteristics of clinical documentation and ontology granularity, rather than extraction errors.

\FloatBarrier

\section{Discussion}
We developed and externally validated RARE-PHENIX, an end-to-end AI framework for rare disease phenotyping that directly models the real-world clinical workflow: extraction of phenotypes from unstructured notes, ontology-grounded standardization, and prioritization of diagnostically informative features. In a large external evaluation of 16{,}357 real-world clinical notes from UDN patients, RARE-PHENIX consistently outperformed a strong deep learning baseline (PhenoBERT \citep{feng2022phenobert}) across ontology-based similarity and precision–recall-F1 metrics. Importantly, the module-based ablation analyses demonstrate that these gains are not attributable to improved extraction alone, but to the combination of ontology-based standardization and prioritization of diagnostically informative phenotypes. Collectively, our findings demonstrate that operationalizing rare disease phenotyping as an end-to-end framework that reflects real-world clinical workflows, rather than as a single extraction task, substantially improves concordance with clinician-curated phenotypes.

Our results highlight several important implications for the design of AI-based phenotyping tools in real-world clinical workflows. First, both RARE-PHENIX and PhenoBERT achieved higher recall than precision, indicating that these methods tend to capture broader phenotype sets than those curated manually. In clinical practice, this tendency is desirable if diagnostically informative phenotypes are surfaced early to minimize the burden of additional clinical review. Our prioritization analysis demonstrates that, compared to random ranking of the same extracted phenotypes, RARE-PHENIX's prioritization approach (Module~3) yields consistent, patient-level improvements across all metrics, with the largest gains observed earlier in the list. This result provides evidence that the prioritization module has the potential to improve the clinical utility of extracted phenotypes. Second, our ablation analysis identified ontology-based standardization as a major contributor to performance gains. Standardizing free-text strings to HPO terms substantially increased precision and F1 while leaving recall largely unchanged. This suggests that a key barrier to effective computational phenotyping in the rare disease setting is normalization into a structured phenotype representation that can support downstream decision-making. Third, the comparable recall achieved by in-context learning with \texttt{ChatGPT-4o} relative to fine-tuned 70b models suggests that in-context learning may be a promising alternative in resource-constrained clinical settings where local model training is not feasible. 

RARE-PHENIX complements existing rare disease LLM-based systems by addressing the bottleneck of extracting diagnostic informative phenotypes directly from clinical notes. Earlier EHR-based approaches primarily focused on retrieval, screening, or rule-based identification of rare disease patients rather than end-to-end phenotyping workflows \citep{garcelon2018clinician, mak2024computer}. Recent studies advanced phenotype extraction through supervised and weakly supervised pipelines, and LLM-based studies focused on optimizing prompting strategies, hybrid dictionary-LLM methods, and ontology grounding to improve extraction accuracy \citep{segura2022exploring, dong2023ontology, cao2024automatic, xiao2023enhancing}. In parallel, many LLM-based rare disease systems prioritize diagnosis or gene and disease ranking, such as automated differential diagnosis pipelines \citep{mao2025phenotype}. By directly modeling the clinical workflow of phenotype extraction, standardization, and diagnostic prioritization, RARE-PHENIX complements existing extraction-based systems and provides a modular phenotyping approach that can enhance downstream diagnostic frameworks, including differential diagnosis systems, phenotype-to-gene prioritization pipelines, and multi-agent diagnostic workflows \citep{mao2025phenotype, zhao2026agentic, kim2024assessing}. 

Our study has several strengths. First, we evaluated RARE-PHENIX using a large and heterogeneous corpus of real-world clinical notes from an external clinical site, enhancing the generalizability of our findings. To our knowledge, this is the first study to operationalize the full clinical workflow of rare disease phenotyping with comprehensive ablation analyses of extraction, standardization, and prioritization modules. Another strength of RARE-PHENIX is its modular design. Because of this flexible design, the extraction and standardization modules can be deployed upstream of other systems that rely on structured HPO inputs. Moreover, the prioritization module can be applied to phenotypes generated by other extraction methods to support diagnostic decision-making. 

Our study also has limitations that should be considered. The UDN cohort is enriched for complex, multi-system cases and may not fully represent general genetics or sub-specialty patient populations with rare diseases. The strongest performance was achieved with large models (i.e., 70b parameters), which may be difficult to deploy in resource-constrained environments. Clinician-curated HPO terms were used as the ground truth in our evaluation; however, manual curation may be incomplete and subject to inter-clinician variability, which may underestimate true model performance for phenotypes present in the clinical notes but not captured in the clinician-curated list. In addition, while RARE-PHENIX achieved favorable performance in retrospective evaluation, its downstream impact on diagnostic accuracy, time to diagnosis, and clinician workload should be evaluated prospectively.

In conclusion, RARE-PHENIX provides a clinically aligned, end-to-end framework for rare disease phenotyping that has the potential to improve the usability of phenotypes extracted from real-world clinical notes. By integrating extraction, ontology-based standardization, and prioritization of diagnostically informative phenotypes into a unified pipeline, the framework has the potential to reduce manual curation burden and serve as a scalable phenotyping tool in rare disease diagnostic workflows. Prospective studies assessing effects on diagnostic efficiency, gene and variant prioritization, and clinician workflow will be essential to establish real-world clinical benefit and to guide responsible deployment in rare disease care settings.

\clearpage
\backmatter

\section*{Acknowledgments}
The authors are grateful to the patients for participating in the Undiagnosed Diseases Network. 

\section*{Data availability}
The Undiagnosed Diseases Network data used in this study contain sensitive patient information. De-identified patient data, including phenotypic and genomic data, are deposited in the \href{https://www.ncbi.nlm.nih.gov/gap/}{database of Genotypes and Phenotypes (dbGaP)} maintained by the National Institutes of Health. To explore data available in the latest release, visit the \href{https://www.ncbi.nlm.nih.gov/projects/gap/cgi-bin/study.cgi?study\_id=phs001232.v7.p3}{UDN study page in dbGaP}. Individuals interested in accessing UDN data through dbGaP should submit a data access request. Detailed instructions for this process can be found on the NIH Scientific Data Sharing website: \href{https://sharing.nih.gov/accessing-data/accessing-genomic-data/how-to-request-and-access-datasets-from-dbgap}{How to Request and Access Datasets from dbGaP}.

\section*{Code availability}
Code used in this study are publicly available at \href{https://github.com/cathyshyr/RARE_PHENIX_for_rare_disease_phenotyping}{\texttt{https://github.com/cathyshyr/RARE\_PHENIX\_for\_rare\_disease\_phenotyping}}.

\section*{Competing interests}
The authors declare no competing interests.

\section*{Funding}
This work was supported in part by the National Institutes of Health Common Fund, grant 15-HG-0130 from the National Human Genome Research Institute, U01NS134349 from the National Institute of Neurological Disorders and Stroke, R00LM014429 from the National Library of Medicine, T32GM082773 from the National Institute of General Medical Sciences, and the Potocsnak Center for Undiagnosed and Rare Disorders.

\section*{Author Contributions}
Concept and design: C.S., L.B., H.X.\\
Drafting of the manuscript: C.S.\\
Critical revision of the manuscript for important intellectual content: C.S., Y.H., R.J.T., T.A.C., K.W.B., R.H., D.V.F., A.W., J.F.P., L.B., H.X.\\
Obtained funding: C.S., R.H.\\
Administrative, technical, or material support: C.S., Y.H., R.J.T., T.A.C., K.W.B., R.H., D.V.F., A.W., J.F.P., L.B., H.X.\\
Supervision: L.B., H.X.

\clearpage
\bibliography{sn-bibliography}

\end{document}